\documentclass{article}

\usepackage[final,main]{neurips_2026}

\usepackage[utf8]{inputenc}
\usepackage[T1]{fontenc}
\usepackage{hyperref}
\usepackage{url}
\usepackage{booktabs}
\usepackage{amsfonts}
\usepackage{amsmath}
\usepackage{nicefrac}
\usepackage{microtype}

\usepackage[table]{xcolor}
\usepackage{graphicx}
\usepackage{multirow}

\usepackage{gensymb}
\usepackage{amssymb}
\newcommand{\pano}{PanoWorld}
\newcommand{\dataset}{PanoGeo}

\definecolor{lightblue}{RGB}{173,216,230}
\usepackage{wrapfig} 
\newcommand{\intro}{
\begin{figure*}[h]
\centering
\includegraphics[width=\textwidth]{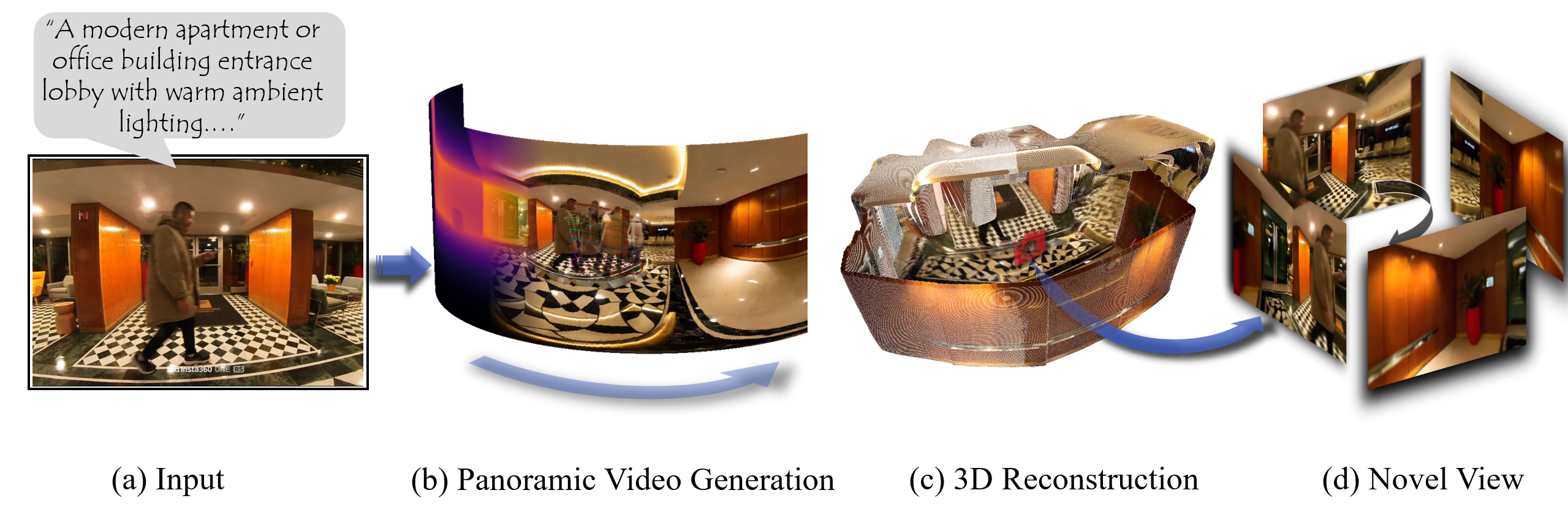}
\vspace{-0.2in}
\caption{\textbf{PanoWorld} generates geometry-consistent $360^{\circ}$ panoramic video from a single perspective image and text prompt. Unlike prior panoramic video methods that optimize for visual realism alone, PanoWorld enforces depth and trajectory consistency in the latent world state, so that its output can also be lifted to 3D (qualitative illustration): (a) input perspective image, (b) generated $360^{\circ}$ panoramic video, (c) point cloud lifted from the generated video, and (d) rendering from a nearby camera pose unseen during generation.}
\label{fig:intro}

\end{figure*}

}

\newcommand{\pipeline}{
\begin{figure*}[t]
\centering
\includegraphics[width=\textwidth]{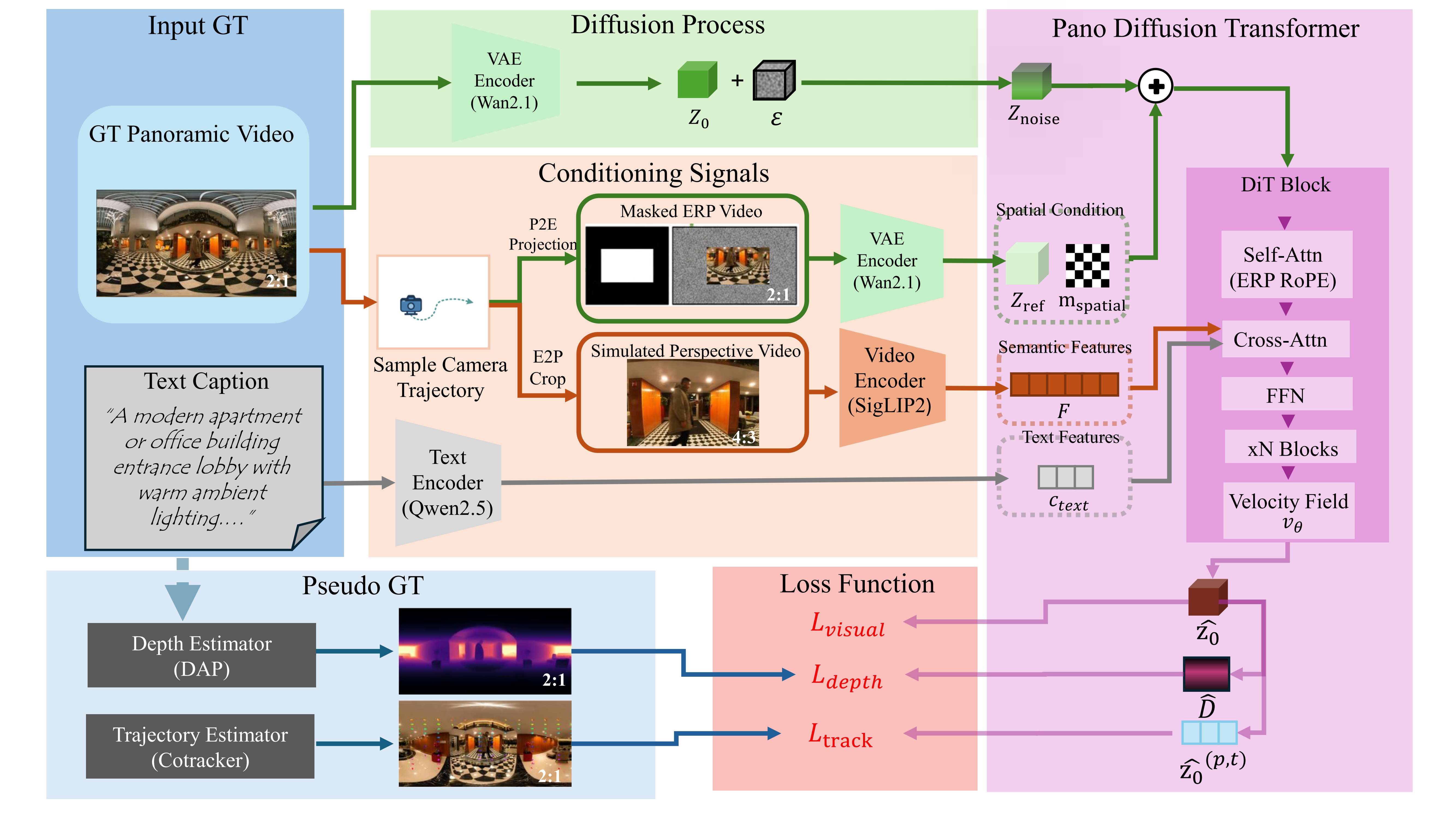}
\vspace{-0.2in}
\caption{\textbf{Training pipeline overview.}
A ground-truth panoramic video from \dataset{} (\textcolor{blue!70!black}{blue, left}) is
VAE-encoded into a clean latent $\mathbf{z}_0$ and mixed with noise $\boldsymbol{\epsilon}$ to
form $\mathbf{z}_{\text{noise}}$ (\textcolor{green!50!black}{green, top}).
Three conditioning signals (\textcolor{orange!80!black}{orange, center}) are derived from a
randomly sampled perspective crop:
(1)~\emph{spatial}: the crop is projected onto the equirectangular canvas (P2E) and VAE-encoded
into reference latent $\mathbf{z}_{\text{ref}}$ and mask $\mathbf{m}_{\text{spatial}}$,
concatenated with $\mathbf{z}_{\text{noise}}$.
(2)~\emph{semantic}: frozen SigLIP2 encodes the crop into per-frame features $\mathbf{F}$,
injected via cross-attention.
(3)~\emph{text}: caption features $\mathbf{c}_{\text{text}}$, also via cross-attention.
The Pano Diffusion Transformer (\textcolor{purple!70!black}{purple, right}) processes
$\mathbf{z}_{\text{input}} {=} [\mathbf{z}_{\text{noise}};\mathbf{z}_{\text{ref}};\mathbf{m}_{\text{spatial}}]$
through $N$ DiT blocks with ERP-aware RoPE and predicts velocity $v_\theta$, from which
$\hat{\mathbf{z}}_0$ is recovered.
\textcolor{red}{Three losses} supervise $\hat{\mathbf{z}}_0$:
$\mathcal{L}_{\text{visual}}$ (rectified-flow target),
$\mathcal{L}_{\text{depth}}$ (Eq.~\ref{eq:depth_loss}, depth head vs.\ Depth Any Panoramas (DAP) pseudo-labels~\citep{lin2025depth}),
and $\mathcal{L}_{\text{track}}$ (Eq.~\ref{eq:track_loss}, 3D point consistency of
CoTracker3~\citep{karaev2024cotracker} tracks lifted via the predicted depth).
All pseudo-labels (\textcolor{blue!70!black}{blue, bottom left}) are precomputed offline at
zero training-time cost.
See Sec.~\ref{sec:method} for details.}
\label{fig:pipeline}
\vspace{-0.2in}
\end{figure*}
}

\newcommand{\qualit}{
\begin{figure*}[t]
\centering
\includegraphics[width=\textwidth]{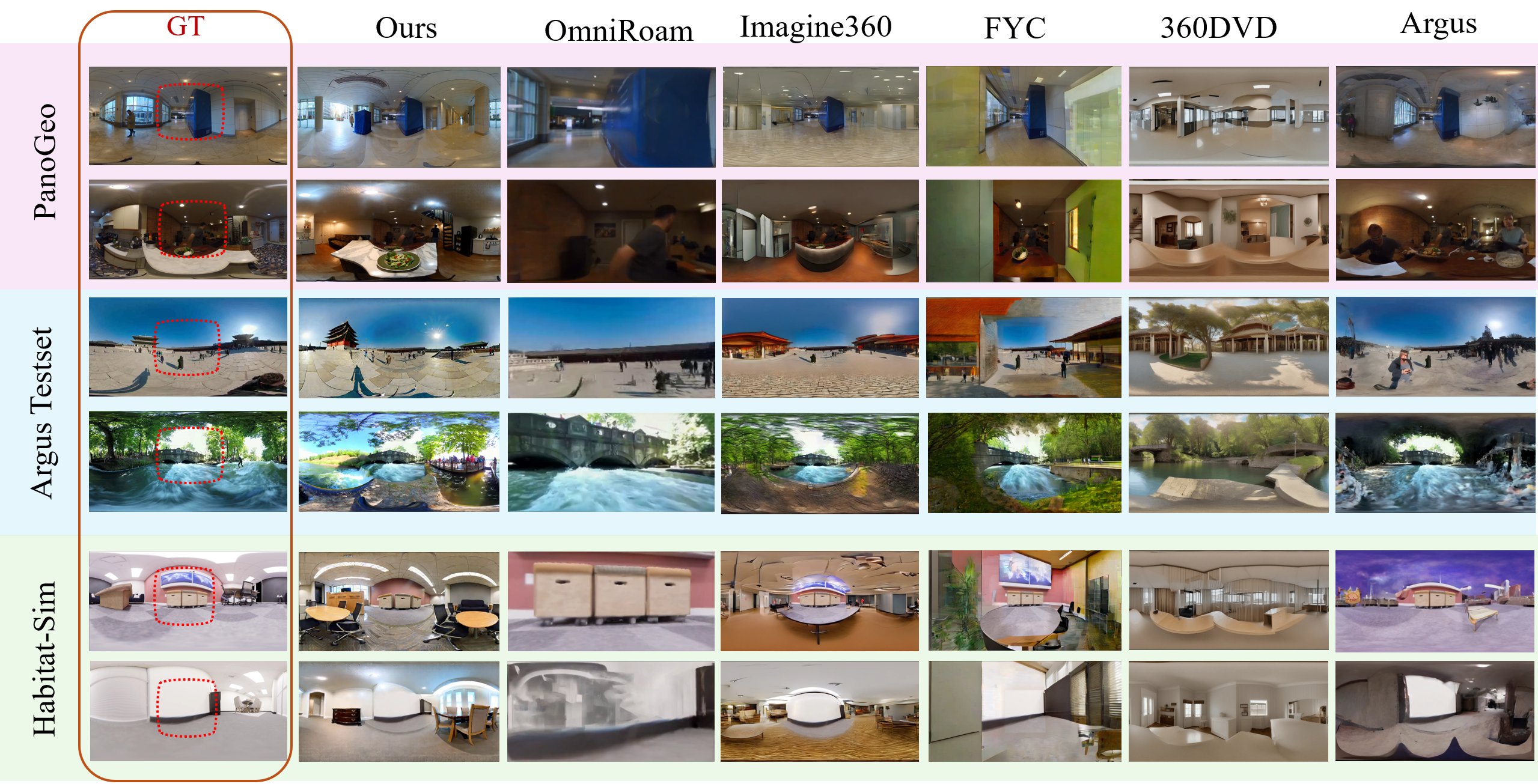}
\vspace{-0.1in}
\caption{\textbf{Qualitative comparison on \dataset{} under Stage~1 input (single perspective image $+$ caption).}
One representative ERP frame per method on six clips spanning the three evaluation regimes: PanoGeo (rows~1--2), Argus Testset (rows~3--4), and Habitat-Sim (rows~5--6).
The \textcolor{red}{red dashed box} on each GT frame marks the perspective crop fed to all methods; the rest of the sphere is hallucinated.
Compared methods: OmniRoam~\citep{liu2026omniroam}, Imagine360~\citep{tan2025imagine360}, Follow-Your-Canvas (FYC)~\citep{chen2025infinite}, 360DVD~\citep{wang2024360dvd}, and ARGUS~\citep{luo2025argus}, all run under the input-parity protocol of Sec.~\ref{sec:exp_setup}.
Stage~2 (ERP first-frame input) is omitted from this figure as a still frame cannot reveal its temporal-anchoring advantage; quantitative Stage~2 results are reported in Table~\ref{tab:main}.}
\label{fig:qualitative}
\vspace{-0.2in}
\end{figure*}

}

\newcommand{\promptfig}{
\begin{figure*}[t]
\centering
\includegraphics[width=\textwidth]{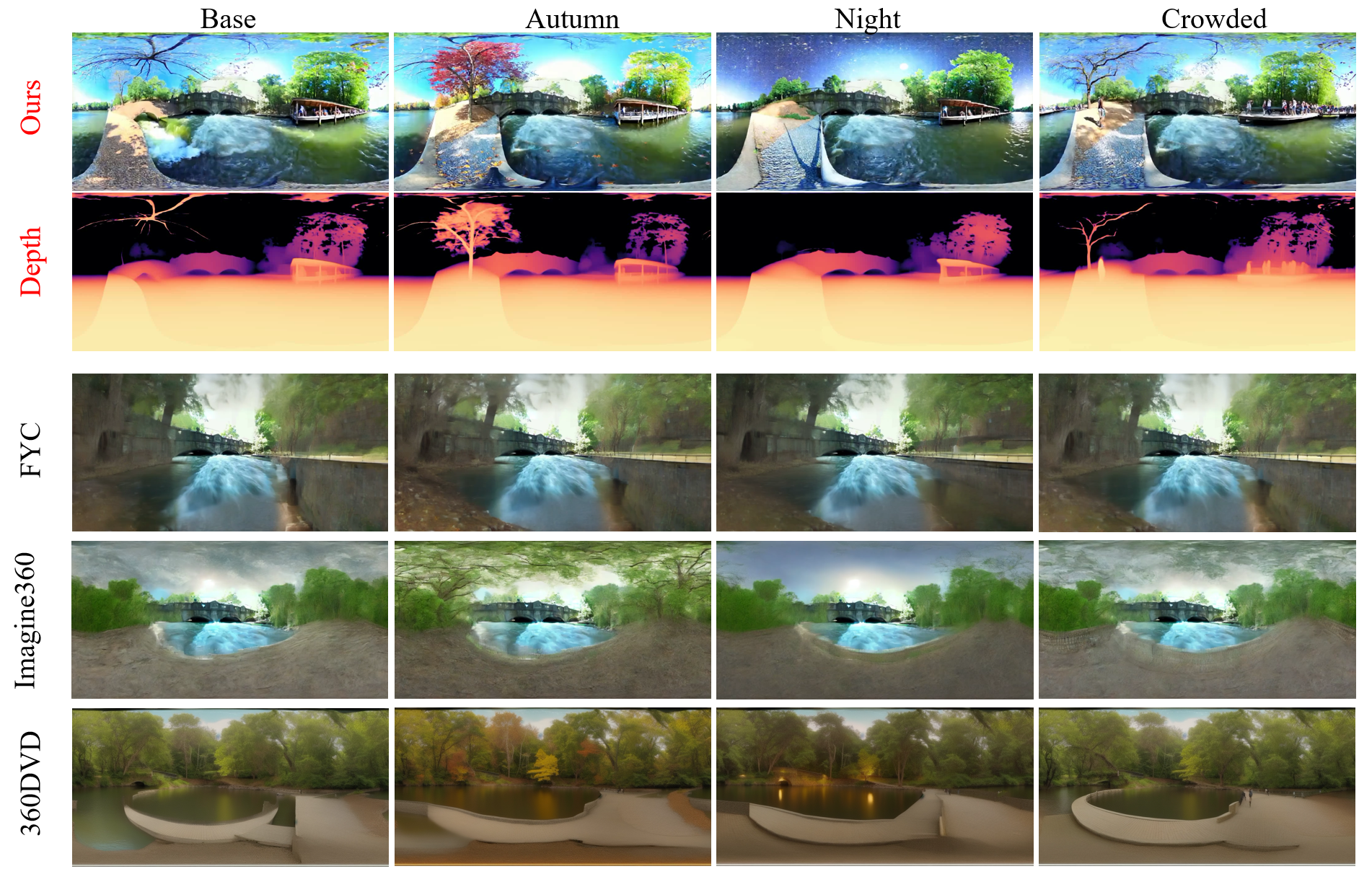}
\vspace{-0.2in}
\caption{\textbf{Prompt controllability from a fixed input image.}
Same input image (an Argus testset clip) and seed; only the last clause of the prompt changes per column. The depth row is DAP on our generated frames (bright is near). ARGUS and OmniRoam take no caption and are omitted.}
\label{fig:prompt_control}
\vspace{-0.2in}
\end{figure*}
}
\usepackage{amssymb} 
\usepackage{pifont}  
\usepackage{xcolor}  
 \usepackage{multirow} 
 \usepackage{float}

\newcommand{\xmark}{\textcolor{red}{\ding{55}}}       

\definecolor{darkgreen}{rgb}{0.0, 0.5, 0.0}

\newcommand{\maintab}{
\begin{table*}[t]
\centering
\small
\caption{\textbf{Quantitative comparison across PanoGeo / Argus testset / Habitat-Sim ($150$ clips, balanced $50/50/50$).}
We report distributional visual realism (FVD, FAED, FID), caption alignment (CLIP-T), and three correspondence-free geometric self-consistency scores computed on the predicted video alone via PaGeR depth and BootsTAPIR 3D tracks (3D-Smooth, Depth-$\sigma$, Tr-Life). Formal definitions and equations for all metrics are in Appendix~\ref{subsec:metric_defs}.
\emph{Top -- Stage~1}: single perspective image $+$ caption input.
\emph{Bottom -- Stage~2}: single ERP first frame $+$ caption (frame-$0$ anchored).
Base models: 360DVD / Imagine360 / Follow-Your-Canvas use AnimateDiff, ARGUS uses SVD, OmniRoam uses WAN~2.1, and \pano{} uses Cosmos~2.5.
Best in \textbf{bold}, second \underline{underlined}.}
\vspace{0.1in}
\label{tab:main}
\setlength{\tabcolsep}{4pt}
\resizebox{\textwidth}{!}{%
\begin{tabular}{l ccc c ccc}
\toprule
\multirow{2}{*}{Method} & \multicolumn{3}{c}{Visual Realism} & {Caption} & \multicolumn{3}{c}{Geometric Self-Consistency} \\
\cmidrule(lr){2-4} \cmidrule(lr){5-5} \cmidrule(lr){6-8}
& FVD$\downarrow$ & FAED$\downarrow$ & FID$\downarrow$ & CLIP-T$\uparrow$ & 3D-Smooth$\downarrow$ & Depth-$\sigma\downarrow$ & Tr-Life$\uparrow$ \\
\midrule
\multicolumn{8}{l}{\emph{Stage~1: single perspective image $+$ caption}} \\
\midrule
360DVD~\citep{wang2024360dvd}                  & 100.7 &  94.6 & 191.8 & 0.262 & 0.763 & 0.065 & 0.881 \\
Imagine360~\citep{tan2025imagine360}         &  67.7 &  96.6 & 142.9 & 0.253 & 0.364 & 0.099 & 0.594 \\
ARGUS~\citep{luo2025argus}                   &  65.0 & \textbf{82.7} & 152.5 & 0.212 & \textbf{0.100} & \underline{0.053} & \textbf{0.991} \\
Follow-Your-Canvas~\citep{chen2025infinite}  &  82.2 & 108.8 & 182.3 & \textbf{0.279} & 0.463 & 0.140 & 0.600 \\
OmniRoam~\citep{liu2026omniroam}             &  98.8 & 131.2 & 178.4 & 0.132 & 1.063 & 0.280 & 0.200 \\
\rowcolor{lightblue}\textbf{\pano (Ours)}    & \textbf{56.1} & \underline{83.4} & \textbf{136.4} & \underline{0.264} & \underline{0.101} & \textbf{0.033} & \underline{0.977} \\
\midrule
\multicolumn{8}{l}{\emph{Stage~2: single ERP first frame $+$ caption (frame-$0$ anchored)}} \\
\midrule
OmniRoam~\citep{liu2026omniroam}             & \underline{39.7} & \underline{53.6} & \underline{69.7} & \underline{0.233} & \underline{1.197} & \underline{0.180} & \underline{0.608} \\
\rowcolor{lightblue}\textbf{\pano (Ours)}    & \textbf{28.5} & \textbf{41.5} & \textbf{55.4} & \textbf{0.253} & \textbf{0.511} & \textbf{0.060} & \textbf{0.776} \\
\bottomrule
\end{tabular}%
}
\vspace{-0.2in}
\end{table*}
}

\newcommand{\ablationtab}{
\begin{table}[t]
\centering
\small
\caption{All four variants share the same architecture (ERP RoPE $+$ SigLIP2 $+$ panorama-shaped multi-path conditioning) and the same two-stage training schedule on \dataset{}; only the auxiliary loss weights $\lambda_d, \lambda_\tau$ in Eq.~\ref{eq:total_loss} are toggled. We report the same Stage~1 evaluation regime and the same $150$-clip split as Table~\ref{tab:main}. Best in \textbf{bold}.}
\label{tab:ablation}
\setlength{\tabcolsep}{4pt}
\begin{tabular}{cc cc ccc}
\toprule
\multirow{2}{*}{$\mathcal{L}_{\text{depth}}$} & \multirow{2}{*}{$\mathcal{L}_{\text{track}}$} & \multicolumn{2}{c}{Visual / Caption} & \multicolumn{3}{c}{Geometric Self-Consistency} \\
\cmidrule(lr){3-4} \cmidrule(lr){5-7}
& & FVD$\downarrow$ & CLIP-T$\uparrow$ & 3D-Smooth$\downarrow$ & Depth-$\sigma\downarrow$ & Tr-Life$\uparrow$ \\
\midrule
    \xmark        &      \xmark        & 56.72 & 0.263 & 0.177 & 0.055 & 0.899 \\
\checkmark        &      \xmark        & 57.47 & \textbf{0.267} & 0.149 & 0.041 & 0.933 \\
    \xmark        & \checkmark         & 57.05 & 0.263 & 0.115 & 0.039 & 0.940 \\
\rowcolor{gray!15}
\checkmark        & \checkmark         & \textbf{56.12} & 0.264 & \textbf{0.101} & \textbf{0.033} & \textbf{0.977} \\
\bottomrule
\end{tabular}
\vspace{-0.2in}
\end{table}
}

\title{PanoWorld: Geometry-Consistent Panoramic Video World Modeling}

\author{%
  Le Jiang \And
  Xiangyu Bai \And
  Bishoy Galoaa \And
  Shayda Moezzi \And
  Caleb James Lee \And
  Tooba Imtiaz \And
  Edmund Yeh \And
  Jennifer Dy \And
  Yanzhi Wang \And
  Sarah Ostadabbas$^\dagger$ \\[4pt]
  \texttt{\{jiang.l, bai.xiang, galoaa.b, moezzi.s, lee.calebj,}\\
  \texttt{imtiaz.t, e.yeh, j.dy, yanzhiwang, s.ostadabbas\}@northeastern.edu}\\[2pt]
  \textit{Northeastern University}
}

\begin{document}

\maketitle
\hypersetup{pageanchor=false}
\renewcommand{\thefootnote}{$\dagger$}
\footnotetext{Corresponding author: \texttt{s.ostadabbas@northeastern.edu}}
\renewcommand{\thefootnote}{\arabic{footnote}}
\intro
\begin{abstract}
We present \textbf{\pano}, a panoramic video world model that generates geometry-consistent 360$\degree$ video from a single image and a caption. Existing panoramic video methods optimize primarily for visual realism and do not explicitly constrain the underlying 3D scene state, producing outputs that appear plausible yet exhibit inconsistent depth, broken correspondences, and implausible motion across the spherical surface. We address this gap by framing panoramic video generation as a geometry- and dynamics-consistent latent state modeling problem rather than pure visual synthesis. Building on a pre-trained perspective video world model, we introduce two lightweight regularizers: a depth consistency loss against pseudo ground-truth panoramic depth, and a trajectory consistency loss that supervises the 3D world-frame positions of tracked points across time. We further apply spherical-geometry-aware adaptation to the conditioning and positional encoding. We additionally introduce \textbf{\dataset}, a unified geometry-aware panoramic video dataset with consistent depth, trajectory, and prompt annotations across diverse real and synthetic sources, used for both training and stratified evaluation. Experiments show that \textbf{\pano} improves geometric consistency over prior panoramic generation methods while maintaining competitive visual realism, suggesting that explicit geometric supervision is a practical step toward panoramic world models for embodied AI applications. Code is available at \url{https://github.com/ostadabbas/PanoWorld}.
\end{abstract}

\section{Introduction}
\label{sec:intro}

Video world models~\citep{agarwal2025cosmos, lecun2022worldmodel} have recently emerged as a promising paradigm for learning structured representations of dynamic physical environments. Rather than modeling video purely as appearance sequences, these systems seek internal state representations that support prediction, simulation, and planning~\citep{hafner2019dream, bruce2024genie, russell2025gaia, zhu2025aether}. By training on large-scale video corpora, such models aim to capture scene dynamics, geometry, and semantics, enabling applications in autonomous driving, robotic control, and video understanding~\citep{huang2025vid2world, zhu2025astra,hong2023cogvideo,chen2025teleworld,chi2025wowworldomniscientworld}. A central question, however, is what meaningfully distinguishes a video world model from a high-capacity video generator.

Conventional video generation models typically learn conditional distributions over pixel sequences, either as joint denoising $p(x_{1:T})$~\citep{blattmann2023svd, bar2024lumiere, zheng2024open, brooks2024sora} or as autoregressive prediction $p(x_{t+1}\mid x_{\leq t})$~\citep{hong2023cogvideo, kondratyuk2023videopoet}. By contrast, video world models can be viewed as learning latent dynamical systems:
\begin{equation}
s_{t+1} = f_{\theta}(s_t, a_t), \quad x_t = g_{\phi}(s_t),
    \label{eqn:latent}
\end{equation}
where $s_t \in \mathbb{R}^d$ denotes a latent scene state encoding geometry, object configurations, and physical factors, $a_t$ denotes the conditioning input (the text prompt in \pano{}), $f_{\theta}$ is a learned transition model, and $g_{\phi}$ maps latent states to observations. Video is then generated through state evolution and rendering rather than direct frame-wise synthesis, and structured latent transitions are particularly important when downstream tasks require persistent spatial understanding rather than short-horizon realism~\citep{wu2024ivideogpt}.

Despite recent progress, many current video world models~\citep{agarwal2025cosmos, blattmann2023svd, hong2023cogvideo, marble2025} still operate in the limited field of view of standard perspective cameras. As a result, the latent state at time $t$ is learned from only a partial observation of the surrounding world:
\begin{equation}
    x_t = g_{\phi}(s_t, \Omega_t), \quad \Omega_t \subset \mathbb{S}^2,
\end{equation}

where $\Omega_t$ denotes the visible camera frustum on the viewing sphere $\mathbb{S}^2$ of all directions. This induces a partially observable setting in which critical scene content may lie outside the current view, complicating tasks that require holistic environmental awareness such as immersive telepresence~\citep{huang2025terra, li2026worldgrow}, navigation~\citep{zeng2025futuresightdrive, li2026omninwm}, and robotic manipulation~\citep{AgiBotWorldTeam2025agibot, routray2025vipra}. Multi-view fusion across narrow-view observations is one workaround~\citep{mildenhall2021nerf, chang2025reconviagen}, but persistent identity, geometry, and motion consistency are difficult to maintain across long-range motion or in a shared coordinate frame~\citep{macario2022comprehensive,qin2023review,gao2022review}.

Panoramic (equirectangular) video offers an attractive alternative by directly modeling the full observation domain ($\Omega_t = \mathbb{S}^2$). A single panoramic frame captures the surrounding 360$\degree$ scene in one global coordinate system, enabling unified reasoning over spatial relationships and motion across all directions. Although panoramic representations introduce projection distortions, they substantially reduce viewpoint truncation and fragmented scene state estimation.

Existing panoramic video methods~\citep{luo2025argus, tan2025imagine360, fang2025panovg, wang2024360dvd} remain primarily optimized for visual realism, minimizing appearance-level losses between generated and ground-truth panoramic sequences without explicitly constraining latent state consistency across time or viewpoints. Generated panoramas thus appear realistic while exhibiting inconsistent depth, broken correspondences across adjacent regions, or implausible motion trajectories on the sphere, failure modes that block full-sphere world modeling for embodied applications.

In this work, we propose \textbf{\pano}, a panoramic video world model that formulates 360$\degree$ video generation as a geometry- and dynamics-consistent state modeling problem rather than pure visual synthesis. Starting from a pre-trained video world model, we explicitly regularize latent state evolution across time and viewpoints using complementary supervisory signals: appearance fidelity, 3D geometric consistency via depth supervision~\citep{lin2025depth}, and motion consistency via trajectory alignment~\citep{karaev2024cotracker}. This shifts panoramic generation from frame matching toward structured world-state prediction in a unified spherical coordinate space. Because depth is supervised jointly with appearance, the generated video can also be lifted to a dense 4D point cloud and rendered from nearby viewpoints, which we show qualitatively in Figure~\ref{fig:intro} and Appendix~\ref{sec:supple_3d_pipeline}.

Our contributions are three-fold:

\begin{itemize}
    \item We introduce \pano, a panoramic world-modeling framework that adapts a pre-trained perspective video world model to full $360\degree$ spherical observations, recovering the panoramic state outside the initial camera frustum from a single partial observation while enforcing geometric and temporal consistency over the generated clip.

    \item We propose two lightweight latent regularizers beyond visual similarity: (i) a \emph{depth consistency loss} that promotes multi-view geometric coherence, and (ii) a \emph{trajectory consistency loss} that encourages temporally plausible motion. Together, these objectives induce a more structured latent space with minimal added computational overhead.

    \item We introduce \dataset{}, a unified geometry-aware panoramic video dataset that mixes self-captured 4K 360$\degree$ video, public panorama datasets ~\citep{wang2024360dvd, chen2024360}, and Habitat-rendered simulations ~\citep{puig2023habitat3} across diverse indoor and outdoor scenes, all processed by a single annotation pipeline that produces consistent depth, dense 2D trajectories with camera-motion compensation, and prompts. The same pipeline also yields a stratified evaluation split spanning in-domain held-out, real out-of-domain, and synthetic out-of-domain regimes (Sec.~\ref{sec:exp_setup}, full construction details in the Appendix~\ref{sec:supple_dataset}).

\end{itemize}

\section{Related Work}
\label{sec:related}

\paragraph{Video generation and world models.}
Diffusion transformers~\citep{peebles2023dit} trained on large-scale video corpora produce realistic motion and appearance~\citep{blattmann2023svd, hong2023cogvideo, kondratyuk2023videopoet, bar2024lumiere, brooks2024sora, zheng2024open}, but high perceptual realism does not imply persistent state modeling and these systems often struggle with long-horizon consistency and physically grounded reasoning~\citep{bai2025newton, bansal2024videophy}. Video world models~\citep{ha2018worldmodels, lecun2022worldmodel, hafner2019dream, hafner2019learning, bruce2024genie, russell2025gaia, zhu2025aether, agarwal2025cosmos} unify generation with structured latent dynamics for prediction, simulation, and planning. A complementary direction targets static 3D scenes via NeRF~\citep{mildenhall2021nerf}, 3D Gaussian splatting~\citep{kerbl20233d}, or generative scene priors~\citep{marble2025, huang2025terra, li2026worldgrow, hunyuanworld2025tencent, wang2026one2scene}, capturing space without time. Most existing video world models still operate from perspective frustums $\Omega_t \subset \mathbb{S}^2$~\citep{zhu2025aether, li2025martian, bardes2024revisiting}, leaving holistic full-sphere reasoning open.

\paragraph{Panoramic video generation.}
Panoramic video exposes the full spherical observation domain $(\Omega_t = \mathbb{S}^2)$ in a single coordinate frame and is naturally suited to immersive telepresence, robotics, and scene-centric world modeling. Recent panoramic generators include 360DVD~\citep{wang2024360dvd} (controllable dual-branch diffusion), ARGUS~\citep{luo2025argus} (perspective-conditioned SVD with reference latents and CLIP features), Imagine360~\citep{tan2025imagine360} (perspective-anchor 360$\degree$ synthesis), and ViewPoint~\citep{fang2025panovg} (motion control). A wide-field outpainting line includes Follow-Your-Canvas~\citep{chen2025infinite} and OmniRoam~\citep{liu2026omniroam}, which couples a perspective generator with a long-horizon trajectory module. These methods mainly optimize appearance metrics such as FID~\citep{heusel2017fid}, FVD~\citep{unterthiner2019fvd}, and LPIPS~\citep{zhang2018lpips}, and the latent geometry and temporal coherence of the full sphere are left unconstrained.

\paragraph{Geometry-aware generative modeling.}
Structural priors have been injected into generative models via conditioning signals (ControlNet~\citep{zhang2023controlnet} and follow-ups~\citep{bai2025dctdm}), correspondence-aware attention~\citep{tang2023mvdiffusion}, epipolar constraints~\citep{watson2024nvsplat}, and spherical convolutions~\citep{coors2018spherenet}. A parallel direction fits explicit 3D / 4D representations~\citep{wu2025cat4d, pan2025diff4splat} or feed-forward reconstructors~\citep{wang2025vggt, tang2025mv}. These approaches target image generation, discrete view synthesis, or post-hoc 3D reconstruction, and do not regularize latent state dynamics during panoramic video rollout. \pano{} bridges this gap by treating panoramic video generation as a state-space modeling problem, combining full-sphere observations, latent temporal dynamics, and geometry-aware regularization in a unified training framework.

\section{Introducing \pano}
\label{sec:method}

\pipeline

We learn a panoramic video world model that predicts full-sphere observations from a partial perspective view while preserving coherent geometry and motion. Building on the formulation in Sec.~\ref{sec:intro}, our approach combines (i) spherical-geometry-aware adaptation of a pretrained perspective video world model, (ii) multi-path conditioning from the partial observation, and (iii) two geometry-aware auxiliary objectives -- a depth consistency regularizer $\mathcal{L}_{\text{depth}}$ and a trajectory consistency regularizer $\mathcal{L}_{\text{track}}$ that augment the rectified-flow visual loss $\mathcal{L}_{\text{visual}}$ via pseudo ground-truth depth and trajectories from \dataset{} (full objective: Sec.~\ref{sec:losses}).

\subsection{Training Pipeline Overview}
\label{sec:overview}

We build on Cosmos~Predict~2.5~\citep{agarwal2025cosmos}, a rectified-flow~\citep{liu2023rectifiedflow} DiT video world model with a WAN~2.1 VAE~\citep{wan2025wan} (encoder $\mathcal{E}$, $8\times$ spatial / $4\times$ temporal compression) and a frozen text encoder; off-the-shelf component versions are listed in Appendix~\ref{sec:external_components}. Rather than training a panoramic model from scratch, our goal is to transfer this perspective prior into the spherical domain with minimal additional supervision.

Figure~\ref{fig:pipeline} illustrates the training pipeline. Given a panoramic video from \dataset{}, we sample a random perspective trajectory (FOV $30^\circ$--$120^\circ$, randomized yaw/pitch) that simulates a conventional camera observing a subset of the scene, turning training into a state-completion problem: from partial perspective evidence the model must reconstruct the full-sphere world state and its evolution. The sampled crop conditions the DiT through three complementary pathways (Sec.~\ref{sec:conditioning}): SigLIP2 cross-attention, an equirectangular reference latent with spatial coverage mask, and masked-blending preservation of observed regions. The clean-latent estimate is supervised by geometry-aware losses (Sec.~\ref{sec:losses}): a lightweight depth head maps it to a depth map against DAP pseudo-labels~\citep{lin2025depth}, and tracked points lifted to 3D through that predicted depth are supervised against precomputed GT 3D positions, jointly constraining depth head and denoiser to produce a temporally coherent scene geometry. Training proceeds progressively at $256{\times}512$ and then $512{\times}1024$, with geometry losses active throughout.

\subsection{Equirectangular Adaptation}
\label{sec:equirect}
Equirectangular projection (ERP) maps longitude $\lambda\in[-\pi,\pi]$ and latitude $\phi\in[-\pi/2,\pi/2]$ on the viewing sphere onto a non-uniform 2D image, with horizontal pixel stretching near the poles proportional to $1/\cos\phi$. Standard linear positional embeddings and pixel-uniform losses therefore distort attention and gradients in ways inconsistent with the underlying sphere. We align the model with spherical geometry through three modifications.

\paragraph{Latitude-aware positional embeddings.}
We replace the linear RoPE coordinate $\text{pos}_h = h$ with a latitude-aware position that compresses indices near the poles, where equal pixel increments correspond to smaller angular displacements on the sphere:
\begin{equation}
  \text{pos}_h
  =
  \frac{H-1}{2}\!\left(\sin\!\left(\frac{\pi h}{H-1}-\frac{\pi}{2}\right)+1\right),
  \qquad
  \text{pos}_w = w.
\end{equation}
Equivalently $\text{pos}_h \propto \sin\phi(h) + 1$ where $\phi(h) = \pi h/(H-1) - \pi/2$ is the latitude of row $h$, matching the true angular spacing of ERP rows. Longitude is left linear since ERP preserves uniform angular spacing along the width axis.

\paragraph{Spherical area weighting.}
Because ERP pixels near the poles represent smaller solid angles on the sphere than equatorial
pixels, unweighted pixel-space losses over-represent polar content during optimization.
We correct this by weighting all pixel-space loss terms by $\cos\phi(h)$, which approximates the
spherical area element $\mathrm{d}\Omega = \cos\phi\,\mathrm{d}\phi\,\mathrm{d}\lambda$.
This weighting is applied to the ERP-pixel-space geometric objectives (Eq.~\ref{eq:depth_loss} and Eq.~\ref{eq:track_loss}) so their gradient contributions are proportional to the surface area each pixel represents. The latent-space rectified-flow objective $\mathcal{L}_{\text{visual}}$ is left unweighted since ERP latitude has no direct counterpart at the compressed latent grid.

\paragraph{Circular shift augmentation.}
The left and right boundaries of an equirectangular image are adjacent in 3D space
($\lambda = -\pi$ and $\lambda = \pi$ are the same meridian), yet a standard spatial transformer
treats them as maximally distant.
We apply random horizontal circular shifts to each training panorama, uniformly sampling a
column offset from $\{0, \ldots, W-1\}$ and rolling the image along the width axis.
This augmentation exposes the model to all possible boundary alignments during training, promoting
seamless wrap-around generation at inference time without boundary artifacts.

\subsection{Video-Conditioned Panoramic Generation}
\label{sec:conditioning}

The conditioning pathways accept a perspective video; a single input image is treated as a static clip.

For semantic conditioning, each observed frame is encoded by a frozen image-level semantic encoder (SigLIP2~\citep{tschannen2025siglip}). The resulting dense spatial tokens are projected into the DiT feature dimension and injected into each block through cross-attention. Compared with conventional CLIP-style encoders~\citep{radford2021clip}, SigLIP2 yields substantially richer spatial tokenization, which we find improves scene and object grounding across time. During training, we apply $10\%$ feature dropout to enable classifier-free guidance at inference.

For geometric conditioning, each perspective frame is projected onto the equirectangular canvas using the corresponding camera parameters (FOV, yaw, pitch). Unobserved regions are filled with scaled Gaussian noise, and the resulting sequence is encoded into reference latents $\mathbf{z}_{\text{ref}}=\mathcal{E}(\cdot)$. These reference latents are concatenated with the noisy latents $\mathbf{z}_{\text{noise}}$ and a per-frame visibility mask $\mathbf{m}_{\text{spatial}}$ along the channel dimension to form the DiT input (Figure~\ref{fig:pipeline}). Because the projection is performed per frame rather than from a single image, the reference latent preserves temporal variation consistent with the observed motion.

During training and at the bootstrap inference stage, observed regions are reinserted at every denoising step through blended diffusion~\citep{avrahami2022blended}. This preserves evidence consistency while requiring the model to infer unseen panoramic content.

The reference latent and SigLIP2 pathways are co-active at every training step. The base Cosmos temporal conditioning channel (which replaces a small leading block of latent frames with GT during denoising) is kept at a low-probability schedule (Sec.~\ref{sec:exp_setup}), so the panoramic pathways dominate the conditioning signal during training.

\subsection{Geometry-Aware Training Objectives}
\label{sec:losses}

Visual realism alone is insufficient for world modeling. Generated panoramas should correspond to a stable 3D scene whose objects evolve smoothly over time. We therefore regularize the denoiser along two complementary axes, a per-frame depth objective and a multi-frame 3D trajectory objective, both operating on the clean-latent estimate $\hat{\mathbf{z}}_0 = \mathbf{z}_\sigma - \sigma\, v_\theta(\mathbf{z}_\sigma, \sigma, \mathbf{c})$, where $\sigma\!\in\![0,1]$ is the rectified-flow noise level, $\mathbf{z}_\sigma=(1-\sigma)\mathbf{z}_0 + \sigma\boldsymbol{\epsilon}$ is the noisy latent, $v_\theta$ is the DiT-parameterized velocity predictor, and $\mathbf{c}=(\mathbf{c}_{\text{sem}}, \mathbf{z}_{\text{ref}}, \mathbf{m}_{\text{spatial}})$ bundles the conditioning context of Sec.~\ref{sec:conditioning}. The main training objective is the standard rectified-flow velocity-matching loss~\citep{liu2023rectifiedflow}, $\mathcal{L}_{\text{visual}} = \mathbb{E}_{\sigma,\mathbf{z}_0,\boldsymbol{\epsilon}}\bigl\|v_\theta(\mathbf{z}_\sigma, \sigma, \mathbf{c}) - (\boldsymbol{\epsilon}-\mathbf{z}_0)\bigr\|_2^2$. The two auxiliary objectives below augment it with geometric supervision on $\hat{\mathbf{z}}_0$ using pseudo-labels precomputed offline on \dataset{}, requiring no external model inference at training time.

Because auxiliary supervision is unreliable when $\hat{\mathbf{z}}_0$ is still noisy, both auxiliary losses are modulated by a noise-adaptive confidence factor $c(\sigma) = \mathbf{1}[\sigma < \sigma_{\max}] \cdot (1 - \sigma/\sigma_{\max})_+^2$ that fades them in quadratically as denoising progresses ($\sigma_{\max}$ in Appendix~\ref{sec:training}). Pixel-space terms are additionally weighted by the spherical area factor $w_h = \cos\phi(h)$ (Sec.~\ref{sec:equirect}).

\paragraph{Depth consistency loss.}
A lightweight depth head (compact 3D conv network, ${\sim}138$K parameters) maps
$\hat{\mathbf{z}}_0$ to a normalized depth map $\hat{D}\in[0,1]^{1\times T'\times H'\times W'}$,
supervised by panoramic depth pseudo-labels $D^{gt}$~\citep{lin2025depth}, which are
computed natively on equirectangular input. Because both $\hat{D}$ and $D^{gt}$ are normalized
to $[0,1]$, scale-invariance is unnecessary, and we adopt an L1 formulation augmented with an
edge-preserving gradient term. Let $\mathcal{M} = \{(t,h,w) : 0.01 < D^{gt}_{t,h,w} < 0.95\}$
mask out saturated sky and near-range artefacts, and let $\mathcal{M}_q \subseteq \mathcal{M}$
further exclude the top $2\%$ per-pixel residuals, which empirically correspond to
pseudo-label annotation outliers. Writing $\nabla_a$ for finite differences along axis
$a\in\{h,w\}$ with neighbour-valid subset $\mathcal{M}_a$, the depth loss is
\begin{equation}
\mathcal{L}_{\text{depth}}
=
c(\sigma)\!\left[
\frac{1}{|\mathcal{M}_q|}\!\!\sum_{(t,h,w)\in\mathcal{M}_q}\!\!\!w_h\,\bigl|\hat{D}_{t,h,w}-D^{gt}_{t,h,w}\bigr|
\;+\;
\frac{1}{2}\!\sum_{a\in\{h,w\}}\!\frac{1}{|\mathcal{M}_a|}\!\sum_{\mathcal{M}_a}\!\bigl|\nabla_a\hat{D}-\nabla_a D^{gt}\bigr|
\right]\!,
\label{eq:depth_loss}
\end{equation}
where the first term encodes absolute depth and the second sharpens discontinuities at object
boundaries. Compared with scale-invariant log loss~\citep{eigen2014depth}, we find direct L1
preferable here because the data is already pre-normalized, and because inverse-depth objectives
conflict with ERP area weighting by over-emphasizing near-range pixels.

\paragraph{Trajectory consistency loss.}
Per-frame depth alone does not enforce temporal identity: the same physical point may
receive inconsistent depth estimates across frames, and dynamic objects may exhibit
physically implausible motion. We therefore lift each tracked 2D point at frame $t$ into a
world-frame 3D position via the predicted depth and supervise it directly against the
precomputed GT 3D track from \dataset{}. For CoTracker3~\citep{karaev2024cotracker} track
$p$ at frame $t$ with normalized ERP coordinate $\mathbf{u}_{p,t}\in[0,1]^2$ and visibility
$v_{p,t}\in\{0,1\}$, we bilinearly sample depth $\hat{D}(\mathbf{u}_{p,t})$ at the
corresponding latent frame, apply the spherical unprojection
$\pi_{\mathbb{S}^2}:[0,1]^2\times[0,1]\!\to\!\mathbb{R}^3$ (which maps a normalized ERP
coordinate and a depth scalar to the corresponding 3D ray endpoint in the camera frame),
and rotate to the world frame via the per-frame camera pose $(\mathbf{R}_t,\mathbf{t}_t)$
(identity for static-camera clips), so that ego-motion in hand-held and walking captures is not supervised as object motion:

\begin{equation}
\mathbf{X}_{p,t}^{\text{pred}}
=
\mathbf{R}_t^{\!\top}\!\bigl(\pi_{\mathbb{S}^2}(\mathbf{u}_{p,t},\,\hat{D}(\mathbf{u}_{p,t}))-\mathbf{t}_t\bigr).
\label{eq:track_lift}
\end{equation}

Beyond absolute position, we additionally supervise the first- and second-order forward differences of the trajectory by stacking position, velocity, and acceleration into an augmented state $\boldsymbol{\xi}_{p,t} = [\mathbf{X}_{p,t};\, \alpha\,\Delta\mathbf{X}_{p,t};\, \beta\,\Delta^2\mathbf{X}_{p,t}]\in\mathbb{R}^9$, where $\Delta, \Delta^2$ denote forward differences along $t$ and $\alpha,\beta$ control the relative weight of velocity and acceleration (Appendix~\ref{sec:training}). Static scene points have zero GT velocity/acceleration, so rigid-scene self-consistency is enforced implicitly while dynamic objects receive non-trivial motion supervision. With visibility- and latitude-weighted gating

$w_{p,t} = v_{p,t}\cdot\cos\phi_{p,t}$ (where $\phi_{p,t}$ is the latitude of
$\mathbf{u}_{p,t}$) and top-$2\%$ trimming over per-element residuals
$\|\boldsymbol{\xi}_{p,t}^{\text{pred}}-\boldsymbol{\xi}_{p,t}^{\text{gt}}\|_1$, the
trajectory loss is
\begin{equation}
\mathcal{L}_{\text{track}}
=
c(\sigma)\,\frac{1}{\sum_{p,t} w_{p,t}}\sum_{p,t}
w_{p,t}\,\bigl\|\boldsymbol{\xi}_{p,t}^{\text{pred}}-\boldsymbol{\xi}_{p,t}^{\text{gt}}\bigr\|_1.
\label{eq:track_loss}
\end{equation}
Lifting to 3D before comparison tightly couples depth and trajectory supervision: drift in either prediction raises the loss, so the depth head and denoiser must jointly produce a temporally coherent scene geometry. The final training objective combines the rectified-flow visual loss with the two geometry-aware regularizers via scalar weights $\lambda_d, \lambda_\tau \in \mathbb{R}_+$,
\begin{equation}
\mathcal{L}
= \mathcal{L}_{\text{visual}}
+ \lambda_d\,\mathcal{L}_{\text{depth}}
+ \lambda_\tau\,\mathcal{L}_{\text{track}},
\label{eq:total_loss}
\end{equation}

with $\lambda_d{=}0.3$ and $\lambda_\tau{=}0.06$ throughout (full schedule in Appendix~\ref{sec:training}). These regularizers are especially important in the panoramic setting, where each frame already covers the full sphere and no direct conditioning remains for the ${>}90\%$ of pixels outside the input frustum: $\mathcal{L}_{\text{depth}}$ stabilizes global 3D structure across the clip while $\mathcal{L}_{\text{track}}$ preserves plausible object motion across previously unseen panoramic regions.

\section{Experiments}
\label{sec:experiments}
\qualit
\subsection{Experimental Setup}
\label{sec:exp_setup}

\textbf{Dataset.}

We use \dataset{} for both training and evaluation (sources, sizes, and captioning in Appendix~\ref{sec:supple_dataset}). The $150$-clip evaluation suite spans three sources that probe distinct generalization axes: \dataset{} ($50$, held-out from our self-captured corpus), the test split of ARGUS~\citep{luo2025argus} ($50$), and Habitat-Sim ~\citep{puig2023habitat3} ($50$, synthetic); aggregate scores weight the three sources equally. The self-captured corpus is acquired in publicly accessible spaces; all faces are anonymized via Deface~\citep{xu2020centerface} prior to training and evaluation.

\textbf{Implementation.}
We train \pano{} in two progressive stages: Stage~1 at $256 \times 512$ for $6{,}000$ iterations and Stage~2 at $512 \times 1024$ for $2{,}000$ iterations. Full optimization, loss hyperparameters, and the two-stage inference pipeline are detailed in Appendix~\ref{sec:training}~and~\ref{sec:inference}.

\textbf{Evaluation.}
We compare \pano{} against 360DVD~\citep{wang2024360dvd}, Imagine360~\citep{tan2025imagine360}, ARGUS~\citep{luo2025argus}, Follow-Your-Canvas~\citep{chen2025infinite}, and OmniRoam~\citep{liu2026omniroam} under two input regimes per clip: \emph{Stage~1} (single perspective image $+$ caption) and \emph{Stage~2} (single ERP first frame $+$ caption, frame-$0$ anchored). Because Stage~1 induces per-method camera paths that misalign with GT, we report quality along two correspondence-free axes that are well-defined regardless of the predicted trajectory: distributional visual realism (FVD~\citep{unterthiner2019fvd}, FAED, FID~\citep{heusel2017fid}), caption alignment (CLIP-T~\citep{radford2021clip}), and geometric self-consistency computed on the predicted video alone via PaGeR~\citep{bozic2026unified} and BootsTAPIR~\citep{doersch2024bootstap}, independent of our DAP/CoTracker3 training labels (3D-Smooth, Depth-$\sigma$, Tr-Life; DAP/CoTracker3 scores in Appendix~\ref{subsec:evaluator_check}). The input-parity protocol and metric equations are in Appendix~\ref{sec:supple_eval}.

\subsection{Comparison with Baselines}
\label{sec:main_results}
\maintab
Table~\ref{tab:main} reports Stage~1 (top) and Stage~2 (bottom) results aggregated over the $150$-clip suite; the per-source breakdown is in Table~\ref{tab:per_regime}.

\textbf{Distributional visual realism.}
\pano{} obtains the lowest FVD ($56.1$, $-13.7\%$ vs.\ ARGUS) and the lowest FID ($136.4$ vs.\ Imagine360 $142.9$), confirming that geometry conditioning does \emph{not} sacrifice visual quality. On FAED \pano{} ($83.4$) trails ARGUS ($82.7$) by a thin margin: ARGUS's near-static rollouts match GT temporal statistics easily under a transformer 3D encoder. The two AnimateDiff baselines (360DVD, FYC) and OmniRoam show large gaps on every distributional metric ($+26\!\sim\!45$ FVD, $+42\!\sim\!55$ FID).

\textbf{Caption alignment.}
\pano{} reaches CLIP-T $0.264$, second only to Follow-Your-Canvas ($0.279$). FYC's edge stems from its fixed-FOV outpainting: it never moves the camera away from the prompt-relevant input crop, while full-sphere methods must distribute caption-relevant content across the entire panorama. \pano{} still substantially outperforms all five-DoF baselines (ARGUS $0.212$, OmniRoam $0.132$).

\textbf{Geometric self-consistency.}
The geometry losses translate directly into the three self-consistency scores, the central claim of this work. \pano{} attains the lowest Depth-$\sigma$ ($0.033$, $-37\%$ vs.\ ARGUS) and is a close second on 3D-Smooth ($0.101$ vs.\ $0.100$) and Tr-Life. The contrast is starkest against OmniRoam, whose unconstrained backbone yields a 3D-Smooth an order of magnitude worse ($1.06$) and tracks that survive only $20\%$ of frames -- evidence that purely visual objectives admit large geometric drift even when frames look plausible. The AnimateDiff baselines (360DVD, Imagine360, FYC) sit $3.6$--$7.6\times$ worse on 3D-Smooth, consistent with per-frame ControlNet-style conditioning. ARGUS needs video input, so it receives a frame-replicated static clip whose quasi-static rollouts favor smoothness; its 3D-Smooth lead stems from its own test split ($0.175$ vs.\ $0.216$), while \pano{} is lower on \dataset{} ($0.052$ vs.\ $0.071$) and Habitat-Sim ($0.035$ vs.\ $0.054$; Table~\ref{tab:per_regime}).

\textbf{Stage~2 (ERP first frame anchored).}
In the Stage2 of Table~\ref{tab:main}, \pano{} dominates OmniRoam on all $7$ metrics ($-28\%$ FVD, $-23\%$ FAED, $-21\%$ FID, $+9\%$ CLIP-T, $-57\%$ 3D-Smooth, $-67\%$ Depth-$\sigma$, $+28\%$ Tr-Life) with fixed panoramic frame-$0$. \pano{}'s own Stage~$1\!\to\!$Stage~$2$ gap shows the expected behavior: the real ERP anchor sharply lowers FVD ($56.1\!\to\!28.5$), while 3D-Smooth and Tr-Life soften because the ERP-conditioned distribution induces larger learned camera motion.
\promptfig

\textbf{Qualitative Results.}
Figure~\ref{fig:qualitative} compares Stage~1 generations on six clips spanning the three sources. Failure modes cluster by input modality: OmniRoam (trajectory-conditioned, no perspective-to-ERP path) generates a sphere disjoint from the GT layout; 360DVD (caption-only) cannot recover the specific details only based on caption; Imagine360 and ARGUS preserve the perspective region but synthesize the rest of the sphere with weak global structure -- duplicated objects across the antipode and high-frequency noise on fine details. \pano{} preserves the input region faithfully and extends it into a globally coherent panorama, matching the GT layout most closely across all three sources.

\subsection{Prompt Controllability}
\label{sec:prompt_control}

To isolate what the caption controls, we fix the input image and the seed and vary only the text. Each prompt is a fixed \emph{layout prompt}, which describes what lies in the front, right, behind and left of the camera, followed by one \emph{swappable clause} that changes the season, illumination, weather and crowd density (Fig.~\ref{fig:prompt_control}). \pano{} renders each clause where it belongs: a red-and-gold tree appears for Autumn, a starry, moonlit sky replaces the daytime sky for Night, and onlookers fill the bank and the platform for Crowded, while the bridge, river channel and platform stay fixed. The depth maps show the new tree and people entering as near-field geometry rather than as texture painted over the old scene. Among the baselines, ARGUS and OmniRoam take no caption, FYC barely changes across prompts, Imagine360 picks up only the global mood, and the text-only 360DVD follows the clause but cannot keep the input view.

\subsection{Ablation Study}
\label{sec:ablation}
\ablationtab
Table~\ref{tab:ablation} toggles $\lambda_d$ and $\lambda_\tau$ (Eq.~\ref{eq:total_loss}) with everything else fixed. The no-loss baseline (row~$1$) already inherits the Cosmos perspective prior and our panorama-shaped conditioning, so its FVD ($56.7$) is close to the full model; the ablation thus exposes where the two losses move \emph{geometric} stability rather than appearance. $\mathcal{L}_{\text{depth}}$ alone (row~$2$) cuts Depth-$\sigma$ by $24\%$ and 3D-Smooth by $16\%$ while leaving FVD essentially unchanged, evidencing that the depth prior acts as geometric ``free supervision'' rather than a quality regularizer. $\mathcal{L}_{\text{track}}$ alone (row~$3$) is the single most effective change for the trajectory metrics (3D-Smooth $0.177\!\to\!0.115$, Tr-Life $0.899\!\to\!0.940$); it also lowers Depth-$\sigma$ to $0.039$ because lifting 2D tracks through the same depth head back-propagates a depth-style gradient, showing the two losses are mutually reinforcing rather than orthogonal. Stacking both (row~$4$, full model) is best on $4/5$ metrics and cuts 3D-Smooth by $43\%$ ($0.177\!\to\!0.101$) and Depth-$\sigma$ by $39\%$ ($0.055\!\to\!0.033$) versus the no-loss baseline while marginally improving FVD; the ordering is unchanged under DAP/CoTracker3 (Appendix~\ref{subsec:evaluator_check}).

\section{Conclusion}
\label{sec:discussion}

Existing panoramic video generators treat the task as visual synthesis, but downstream embodied applications require a coherent \emph{geometric} state. Our results expose this gap quantitatively: prior baselines remain competitive on distributional realism (FVD/FAED/FID) yet drift on geometric self-consistency (3D-Smooth, Depth-$\sigma$, Tr-Life), indicating visual plausibility without a stable underlying scene. \pano{} closes this gap by jointly enforcing $\mathcal{L}_{\text{depth}}$ and $\mathcal{L}_{\text{track}}$, which couple per-frame depth to temporal point identity and reduce depth and trajectory drift over the clip. Combined with \dataset{} and a progressive two-stage training schedule, this suggests that geometric supervision is a practical step toward world models with full-sphere spatial understanding for embodied AI.

\textbf{Limitations.} Like its Cosmos base model and all compared methods, \pano{} generates fixed $93$-frame clips; arbitrarily long rollouts accumulate drift that we leave to future work. The monocular depth pseudo-labels and the perspective video prior of the base Cosmos model also introduce supervisory bias on equirectangular content.


\newpage
\bibliographystyle{plainnat}
\bibliography{references}
\newpage
\appendix
\section{\dataset{} Dataset Construction}
\label{sec:supple_dataset}

\dataset{} is built around a single \emph{unified} panoramic video annotation pipeline applied to three heterogeneous video sources, so that all training and evaluation clips share identical caption / depth / trajectory pseudo-labels regardless of origin.

\subsection{Video Sources}

\textbf{Self-captured 4K panoramas.}
We capture $4$K ($3840 \times 1920$) $360\degree$ videos with a consumer panoramic camera across diverse environments including hotel lobbies, residential interiors, retail and commercial spaces, museums, urban streets, and natural landscapes. Faces are blurred with Deface~\citep{xu2020centerface}.
From each of the ${\sim}200$ recordings we extract ten-second clips at varying temporal offsets, yielding ${\sim}5{,}000$ clips that capture diverse motion patterns and lighting conditions within each scene.

\textbf{Public panorama datasets.}
We supplement self-captured data with ${\sim}12{,}000$ ERP clips cut from ${\sim}230$ x360~\citep{chen2024360} recordings and ${\sim}2{,}000$ clips from 360DVD~\citep{wang2024360dvd}, providing additional motion categories and outdoor real-world diversity. The model reported in the main paper is trained on a subset of these clips. Clips are re-encoded to a unified $H \times W$ resolution and frame rate before downstream processing.

\textbf{Habitat-rendered simulations (Habitat-Sim split).}
For evaluation, we additionally render $50$ panoramic clips with the Habitat simulator. Each clip uses a randomized camera trajectory inside a sampled scene and is rendered to ERP via a six-face cubemap composite with the same resolution / frame rate as the real-world clips, then passed through the same annotation pipeline. These provide a controlled synthetic split with ground-truth depth that we use only for evaluation.

\subsection{Unified Annotation Pipeline}

All clips, regardless of source, pass through the same three-stage annotation pipeline, which is the core of \dataset{}'s contribution:

\textbf{Captioning.}
Each clip is captioned by a vision-language model with a fixed panoramic-aware prompt template, producing scene-content and dynamics descriptions in a uniform style across sources. Training clips are captioned by Gemini~2.5 Flash and evaluation clips by GPT-5 mini; a random subset of captions is audited manually.

\textbf{Depth annotation.}
Per-frame depth maps are produced by Depth Any Panoramas (DAP)~\citep{lin2025depth}, a foundation model designed for equirectangular input. DAP operates natively on panoramic frames without perspective decomposition, producing geometrically consistent depth across the full $360\degree$ field of view. Maps are stored in half-precision (float16) and serve as the pseudo ground truth $D^{gt}$ for the depth consistency loss (Eq.~\ref{eq:depth_loss}).

\textbf{Trajectory annotation.}
Dense 2D point trajectories and per-point visibility masks are extracted with CoTracker3~\citep{karaev2024cotracker} on a regular query grid. For clips with appreciable camera motion, we lift tracked 2D points to $3$D using DAP depth and the equirectangular spherical projection, then estimate per-frame rigid camera transformations via the Umeyama algorithm with RANSAC outlier rejection, separating camera ego-motion from object motion. The resulting world-frame 3D tracks (and their velocities / accelerations) serve as the pseudo ground truth $\boldsymbol{\xi}_{p,t}^{\text{gt}}$ for the trajectory consistency loss (Eq.~\ref{eq:track_loss}).
The full pipeline is implemented as a resumable batch-processing script with multi-GPU parallelism, enabling annotation of new clips at constant per-clip cost.

\subsection{Evaluation Split}

We curate a candidate pool of $150$ clips spanning the three sources (\dataset{} held-out / Argus testset / Habitat-Sim) and report the main paper on a balanced $150$-clip subsample with $50$ clips per source so that aggregate metrics are not dominated by a single source's idiosyncrasies. We tag every candidate clip with three stratification axes: \texttt{scene\_type} $\in$ \{textured\_indoor, outdoor\_with\_sky, mixed\_low\_texture\}, \texttt{motion\_kind} $\in$ \{static, moving\}, and continuous \texttt{sky\_pct} (fraction of sky pixels), and pick clips greedily so each (source, scene\_type) cell receives a representative subset.

\textbf{\dataset{} held-out ($50$ clips).} Held out from our self-captured corpus before training, with the same camera and the same general environment categories but unseen scenes; this is the in-distribution slice of \dataset{}.

\textbf{Argus testset ($50$ clips, sub-sampled from $50$ candidates).} Sampled from the public Argus YouTube test split~\citep{luo2025argus}, covering real-world panoramic content from a different capture distribution (different cameras, different photographers); we subsample to $50$ along the \texttt{motion\_kind} $\times$ \texttt{scene\_type} grid to match the per-source budget of the other two slices.

\textbf{Habitat-Sim ($50$ clips).} Rendered de novo in Habitat with randomized scenes and trajectories, providing a controlled synthetic distribution with deterministic ground-truth depth.

All three sources are evaluated under the identical annotation pipeline and the identical two-stage inference protocol, so per-source metrics differ only because of the underlying video distribution.

\paragraph{Per-source breakdown.}
The aggregate numbers in the main paper (Table~\ref{tab:main}) hide source-level behaviour that is informative for a panoramic generator's failure modes: the \dataset{} held-out slice measures within-distribution polish, the Argus testset stresses appearance priors against unseen real-world capture conditions, and Habitat-Sim stresses geometric priors against synthetic but precisely controlled scenes. Table~\ref{tab:per_regime} reports the same Stage~1 metrics as Table~\ref{tab:main} broken down by source ($50$ clips each).

\begin{table}[t]
\centering
\small
\setlength{\tabcolsep}{3pt}
\caption{\textbf{Per-source Stage~1 breakdown.} Same metric panel as Table~\ref{tab:main}; each source contains $50$ clips of the balanced subset. Best per (source, metric) in \textbf{bold}, second \underline{underlined}.}
\label{tab:per_regime}
\resizebox{\columnwidth}{!}{%
\begin{tabular}{l l ccc c ccc}
\toprule
Source & Method & FVD$\downarrow$ & FAED$\downarrow$ & FID$\downarrow$ & CLIP-T$\uparrow$ & 3D-Sm$\downarrow$ & D-$\sigma\downarrow$ & T-Life$\uparrow$ \\
\midrule
\multirow{6}{*}{\shortstack{\dataset{}\\held-out}}
  & 360DVD~\citep{wang2024360dvd}                  & 136.19 & 132.81 & 266.48 & 0.263 & 0.189 & 0.052 & 0.889 \\
  & Imagine360~\citep{tan2025imagine360}         & 103.08 & 129.68 & 214.58 & 0.266 & 0.278 & 0.102 & 0.580 \\
  & ARGUS~\citep{luo2025argus}                   &  \underline{89.21} & \underline{103.83} & \underline{211.86} & 0.239 & \underline{0.071} & \underline{0.044} & \textbf{0.993} \\
  & Follow-Your-Canvas~\citep{chen2025infinite}  & 128.55 & 121.97 & 243.25 & \textbf{0.302} & 0.224 & 0.167 & 0.573 \\
  & OmniRoam~\citep{liu2026omniroam}             & 166.89 & 194.68 & 253.86 & 0.135 & 0.954 & 0.321 & 0.122 \\
  & \textbf{\pano (Ours)}                         & \textbf{72.98} & \textbf{101.81} & \textbf{181.46} & \underline{0.274} & \textbf{0.052} & \textbf{0.025} & \underline{0.991} \\
\midrule
\multirow{6}{*}{\shortstack{Argus\\testset}}
  & 360DVD~\citep{wang2024360dvd}                  &  90.77 &  86.94 & 191.58 & 0.257 & 1.996 & 0.081 & 0.847 \\
  & Imagine360~\citep{tan2025imagine360}         &  72.53 &  93.68 & \underline{146.99} & 0.267 & 0.642 & 0.085 & 0.671 \\
  & ARGUS~\citep{luo2025argus}                   & \textbf{54.31} & \textbf{85.12} & \textbf{145.84} & 0.241 & \textbf{0.175} & \textbf{0.048} & \textbf{0.996} \\
  & Follow-Your-Canvas~\citep{chen2025infinite}  &  78.46 & 109.83 & 163.78 & \textbf{0.287} & 1.102 & 0.125 & 0.703 \\
  & OmniRoam~\citep{liu2026omniroam}             & 100.56 & 111.18 & 177.17 & 0.169 & 1.874 & 0.275 & 0.310 \\
  & \textbf{\pano (Ours)}                         &  \underline{56.93} &  \underline{86.01} & 150.65 & \underline{0.278} & \underline{0.216} & \underline{0.050} & \underline{0.971} \\
\midrule
\multirow{6}{*}{Habitat-Sim}
  & 360DVD~\citep{wang2024360dvd}                  & 145.02 & 133.26 & 234.17 & \textbf{0.265} & 0.105 & \underline{0.063} & 0.907 \\
  & Imagine360~\citep{tan2025imagine360}         &  \underline{88.08} & 140.73 & \textbf{176.20} & 0.226 & 0.171 & 0.109 & 0.532 \\
  & ARGUS~\citep{luo2025argus}                   & 106.84 & \underline{129.33} & 210.83 & 0.156 & \underline{0.054} & 0.065 & \textbf{0.984} \\
  & Follow-Your-Canvas~\citep{chen2025infinite}  & 100.72 & 159.24 & 237.18 & \underline{0.248} & 0.064 & 0.130 & 0.525 \\
  & OmniRoam~\citep{liu2026omniroam}             & \textbf{79.48} & 165.81 & \underline{188.61} & 0.093 & 0.240 & 0.243 & 0.167 \\
  & \textbf{\pano (Ours)}                         &  94.99 & \textbf{125.09} & 191.34 & 0.240 & \textbf{0.035} & \textbf{0.024} & \underline{0.971} \\
\bottomrule
\end{tabular}%
}
\end{table}

\textbf{Reading the breakdown.}
\pano{} leads the \dataset{} held-out slice on $5/7$ metrics, the source that most directly reflects whether the depth- and trajectory-aware fine-tuning has internalized our capture distribution. On the Argus testset, ARGUS expectedly wins the appearance metrics (FVD/FAED/FID) and all three geometry scores on its own training distribution, with \pano{} second on five of the six. On Habitat-Sim the picture is mixed because synthetic renderings have unfamiliar lighting and texture statistics that hurt every model's distributional metrics, but \pano{} retains the lowest 3D-Smooth and Depth-$\sigma$ by a wide margin. ARGUS's Tr-Life lead on every source reflects its quasi-static rollouts from frame-replicated input (Sec.~\ref{sec:main_results}).

\section{Evaluation Protocol and Metrics}
\label{sec:supple_eval}

We describe here the full input-parity protocol and metric definitions referenced from the main-paper Experiments section.

\subsection{Two-Stage Input Regime}

We benchmark all methods under two input regimes per clip in the \dataset{} evaluation split, designed so that every baseline receives the same total information budget.

\textbf{Stage~1: single perspective image $+$ caption.}

Each method is given a single perspective image (a $480 \times 640$ crop of the GT panorama's first frame with $90^\circ$ horizontal FOV, facing yaw $0^\circ$ and pitch $0^\circ$) plus the medium-length caption. Methods that natively expect a perspective \emph{video} (Imagine360~\citep{tan2025imagine360}, ARGUS~\citep{luo2025argus}, Follow-Your-Canvas~\citep{chen2025infinite}) receive the same image replicated to their native input length. Methods that natively expect a per-frame trajectory (OmniRoam~\citep{liu2026omniroam}) receive a fixed forward trajectory. \pano{} runs its full two-stage inference pipeline (Sec.~\ref{sec:inference}).

\textbf{Stage~2: single ERP first frame $+$ caption (frame-$0$ anchored).}
Each method is given a single equirectangular frame at $t{=}0$ (the GT panorama's first frame) plus the caption. Only OmniRoam and \pano{} natively support ERP image input in this setting. \pano{} runs Round~2 only, with the supplied ERP frame as the anchor.

Both stages are evaluated with the same correspondence-free metric panel (FVD, FAED, FID, CLIP-T, 3D-Smooth, Depth-$\sigma$, Tr-Life; Sec.~\ref{subsec:metric_defs}), so each (method, stage) pair is directly comparable. All metrics are computed on the generated ERP videos; perspective renderings in figures and on the project page are for visualization only.

Each method generates a $T{=}93$-frame ERP video at $512 \times 1024$ and 16~fps. Predictions and GT are temporally resampled to a common $T_{\text{eval}}{=}80$-frame grid for metric computation, and any per-method frame-rate / clip-length differences are absorbed by this resampling. Metrics are reported in aggregate over the balanced $150$-clip subset (Table~\ref{tab:main}) and per source over \dataset{} held-out / Argus testset / Habitat-Sim (Table~\ref{tab:per_regime}).

\subsection{Metric Definitions}
\label{subsec:metric_defs}

We split metrics into two complementary axes. The \emph{correspondence-free} axis is well-defined under any camera path (both stages); the \emph{correspondence-required} axis assumes frame-aligned predictions against GT and is directly meaningful only when frame~$0$ is anchored (Stage~2).

\paragraph{Notation.}
Let $\hat{V} = \{\hat{V}_t\}_{t=1}^{T_{\text{eval}}}$ with $\hat{V}_t \in [0,1]^{H \times W \times 3}$ denote the predicted ERP video resampled to the common $80$-frame grid, and let $V$ denote the GT counterpart. Geometry scores use $T_g{=}20$ frames uniformly subsampled from $\hat{V}$ at $512{\times}1024$. Running PaGeR~\citep{bozic2026unified} on $\hat{V}$ yields a per-frame metric depth map $\hat{D}_t^{\text{re}} \in \mathbb{R}^{H \times W}$, and running BootsTAPIR~\citep{doersch2024bootstap} on $\hat{V}$ (a $30{\times}30$ query grid on frame~$0$, with ERP circular padding) yields $N$ tracked points with 2D positions $\hat{\mathbf{x}}_{i,t} \in \mathbb{R}^2$, visibilities $v_{i,t} \in \{0,1\}$, and 3D positions $\hat{\mathbf{p}}_{i,t} \in \mathbb{R}^3$ obtained by lifting $(\hat{\mathbf{x}}_{i,t}, \hat{D}_t^{\text{re}}(\hat{\mathbf{x}}_{i,t}))$ through the equirectangular spherical unprojection (the same operation used by Eq.~\ref{eq:track_lift}). Let $c$ denote the medium-length caption.

\paragraph{Correspondence-free distributional metrics (FVD, FAED, FID).}
Each metric is a Fr\'echet distance between the empirical Gaussian fit to the encoded predicted videos and to the encoded GT videos:
\begin{equation}
\label{eq:metric_frechet}
d_{F}\bigl(\hat{\mathcal{V}}, \mathcal{V}\bigr) \;=\; \lVert \mu_{\hat{F}} - \mu_{F} \rVert_2^2 + \mathrm{Tr}\!\bigl(\Sigma_{\hat{F}} + \Sigma_{F} - 2 (\Sigma_{\hat{F}} \Sigma_{F})^{1/2}\bigr),
\end{equation}
where $\mu_{\hat F}, \Sigma_{\hat F}$ and $\mu_F, \Sigma_F$ are the empirical mean / covariance of feature embeddings $\{F(\hat V^{(k)})\}_k$ and $\{F(V^{(k)})\}_k$ over the bucket. The choice of encoder $F$ defines the metric: FVD uses a Kinetics-400-pretrained R(2+1)D-18~\citep{unterthiner2019fvd}, FAED uses a Kinetics-400-pretrained Swin3D-T (a transformer-based 3D encoder, complementary to FVD's convolutional inductive bias), and FID~\citep{heusel2017fid} uses InceptionV3 applied per frame and aggregated.

\paragraph{Caption alignment (CLIP-T).}
The mean cosine similarity between the caption embedding and per-frame CLIP image embeddings:
\begin{equation}
\label{eq:metric_clipt}
\mathrm{CLIP\text{-}T}(\hat V, c) \;=\; \frac{1}{T_{\text{eval}}} \sum_{t=1}^{T_{\text{eval}}} \cos\!\bigl(E_I^{\text{CLIP}}(\hat V_t),\, E_T^{\text{CLIP}}(c)\bigr).
\end{equation}

\paragraph{Geometric self-consistency I -- 3D track smoothness (3D-Smooth).}
The median over visible track positions of the discrete second-order temporal derivative of the lifted 3D track, in meters:
\begin{equation}
\label{eq:metric_3dsmooth}
\mathrm{3D\text{-}Smooth}(\hat V) \;=\; \mathrm{median}_{\substack{1 \le i \le N \\ 1 < t < T_g \\ v_{i,t-1}=v_{i,t}=v_{i,t+1}=1}} \bigl\lVert \hat{\mathbf{p}}_{i,t-1} - 2\hat{\mathbf{p}}_{i,t} + \hat{\mathbf{p}}_{i,t+1} \bigr\rVert_2 .
\end{equation}
A static rigid scene with consistent depth has $\mathrm{3D\text{-}Smooth}\!\to\!0$; jittery 3D positions inflate this score regardless of whether they look plausible in 2D.

\paragraph{Geometric self-consistency II -- depth temporal standard deviation (Depth-$\sigma$).}
For each pixel $(u,v)$ that has a finite, positive predicted depth across all frames, define the temporal mean $\bar{D}(u,v) = \frac{1}{T_g} \sum_t \hat D_t^{\text{re}}(u,v)$ and the temporal standard deviation $\sigma_D(u,v) = \sqrt{\frac{1}{T_g} \sum_t (\hat D_t^{\text{re}}(u,v) - \bar D(u,v))^2}$. We then report:
\begin{equation}
\label{eq:metric_depthsigma}
\mathrm{Depth\text{-}\sigma}(\hat V) \;=\; \mathrm{median}_{(u,v) \in \mathcal{V}_D}\, \frac{\sigma_D(u,v)}{\bar D(u,v)},
\end{equation}
where $\mathcal{V}_D$ is the set of pixels with valid depth in every frame. The per-pixel normalisation by mean depth makes the metric invariant to the absolute depth scale and the size of the scene.

\paragraph{Geometric self-consistency III -- track lifetime fraction (Tr-Life).}
The mean (over tracks) of the fraction of frames each track stays visible:
\begin{equation}
\label{eq:metric_trlife}
\mathrm{Tr\text{-}Life}(\hat V) \;=\; \frac{1}{N} \sum_{i=1}^{N} \frac{1}{T_g} \sum_{t=1}^{T_g} v_{i,t} .
\end{equation}
A high value indicates that point identities survive across the rollout, evidencing a coherent underlying scene rather than visually plausible but topologically inconsistent renderings. Combined with 3D-Smooth and Depth-$\sigma$, the three scores triangulate ``does the generator produce a self-consistent 3D scene''.

\subsection{Evaluator Independence}
\label{subsec:evaluator_check}
Our training labels come from DAP and CoTracker3, so scoring with the same models could reward imitation of the labels rather than geometric consistency. The main paper therefore scores geometry with PaGeR (indoor/outdoor metric scale head chosen once per clip by its CLIP router on the GT first frame) and BootsTAPIR; frame sampling, padding, lifting and camera compensation are otherwise unchanged. Table~\ref{tab:evaluator_check} reports the same scores under DAP/CoTracker3: the ablation ordering is identical; against baselines, the swap lets ARGUS edge ahead on 3D-Smooth and Tr-Life in Stage~1, while \pano{} overtakes OmniRoam on Stage-2 3D-Smooth.

\begin{table}[h]
\centering
\small
\setlength{\tabcolsep}{4pt}
\caption{\textbf{Geometry scores under the DAP/CoTracker3 evaluator} (our training-label models), same $150$ clips as Tables~\ref{tab:main} and~\ref{tab:ablation}.}
\label{tab:evaluator_check}
\begin{tabular}{l ccc}
\toprule
Method & 3D-Smooth$\downarrow$ & Depth-$\sigma\downarrow$ & Tr-Life$\uparrow$ \\
\midrule
\multicolumn{4}{l}{\emph{Stage~1}} \\
360DVD             & 0.204 & 0.038 & 0.922 \\
Imagine360         & 0.115 & 0.045 & 0.915 \\
ARGUS              & \underline{0.032} & \underline{0.021} & \underline{0.993} \\
Follow-Your-Canvas & 0.113 & 0.048 & 0.935 \\
OmniRoam           & 4.079 & 0.299 & 0.231 \\
\pano{} (Ours)     & \textbf{0.025} & \textbf{0.013} & \textbf{0.994} \\
\midrule
\multicolumn{4}{l}{\emph{Stage~2}} \\
OmniRoam           & \textbf{0.774} & 0.147 & 0.628 \\
\pano{} (Ours)     & 0.981 & \textbf{0.032} & \textbf{0.896} \\
\midrule
\multicolumn{4}{l}{\emph{Ablation} ($\mathcal{L}_{\text{depth}}$ / $\mathcal{L}_{\text{track}}$)} \\
\xmark\ / \xmark            & 0.051 & 0.028 & 0.958 \\
\checkmark\ / \xmark        & 0.041 & 0.016 & 0.981 \\
\xmark\ / \checkmark        & 0.031 & 0.014 & 0.992 \\
\checkmark\ / \checkmark    & \textbf{0.025} & \textbf{0.013} & \textbf{0.994} \\
\bottomrule
\end{tabular}
\end{table}

\section{External Components}
\label{sec:external_components}

The main paper refers to the off-the-shelf components below by their model names. This section lists their exact public versions for reproducibility.
The base video world model is Cosmos~Predict~2.5~\citep{agarwal2025cosmos} (2B post-trained checkpoint), which bundles the WAN~2.1 VAE~\citep{wan2025wan} as the spatiotemporal tokenizer and Qwen2.5-VL-7B as the text encoder.
For semantic conditioning we use SigLIP2~So400m/14 at $384^2$ input resolution (1152-d hidden, 729 tokens per frame), which is referred to in the main paper simply as \emph{SigLIP2}~\citep{tschannen2025siglip}.
Pseudo-label annotations on \dataset{} are produced once offline using DAP~\citep{lin2025depth} for per-frame equirectangular depth and CoTracker3~\citep{karaev2024cotracker} for dense 2D point tracks. Both are kept frozen throughout training and inference.
Geometry metrics are scored with independent models: PaGeR~\citep{bozic2026unified} (unified \texttt{prs-eth/PaGeR} checkpoint) and BootsTAPIR~\citep{doersch2024bootstap} (PyTorch \texttt{bootstapir\_checkpoint\_v2}).
The metric encoders used for FVD / FAED / FID / CLIP-T are described in Sec.~\ref{sec:supple_eval}.

\section{Two-Stage Training Strategy}
\label{sec:training}

We adopt progressive fine-tuning from the base Cosmos 2B pre-trained weights, with the SigLIP2 projection layer ($\mathbf{W}_{\text{proj}} \in \mathbb{R}^{1152 \times 2048}$) and I2V cross-attention layers randomly initialized:

\textbf{Stage 1: Low-resolution structure learning.}
Training at $256 \times 512$ resolution with 93 frames ($T' = 24$ latent frames) for 6,000 iterations.
The reduced resolution enables faster iteration on a single H100 80GB GPU, allowing the model to learn panoramic spatial structure, horizontal wrap-around behavior, SigLIP2 feature mapping, and the blended diffusion conditioning mechanism.
The model is trained from the base Cosmos 2B pre-trained weights (not from the CLIP-based v3 checkpoint), since the SigLIP2 projection layer requires fresh learning of the new feature space.

The model is trained from the base Cosmos 2B pre-trained weights, and the SigLIP2 projection layer is learned from scratch.

\textbf{Stage 2: High-resolution quality refinement.}
Training at $512 \times 1024$ resolution with 93 frames, initialized from the Stage 1 checkpoint, for 2,000 iterations.
This stage refines texture detail and generation quality while building on the spatial foundations from Stage 1.

Both stages use random FOV sampling ($30^\circ\sim120^\circ$), random yaw rotation, and horizontal continuity augmentation. The base Cosmos temporal conditioning channel is kept at the standard schedule ($p(\text{num\_cond\_frames}{\in}\{0,1,2\}){=}\{0.5,0.25,0.25\}$), with the SigLIP2 cross-attention and reference-latent pathways always co-active and supplying the dominant panoramic conditioning at every training step.

\textbf{Optimizer and loss hyperparameters.}
Both stages use AdamW ($\beta_1{=}0.9$, $\beta_2{=}0.999$, base learning rate $2^{-15}{\approx}3.05{\times}10^{-5}$ under the base Cosmos warm-up and linear-decay schedule, weight decay $10^{-3}$, batch size $1$) on $T{=}93$-frame clips at 16~fps. The geometry loss weights are $\lambda_d{=}0.3$ and $\lambda_\tau{=}0.06$ in $\mathcal{L} = \mathcal{L}_{\text{visual}} + \lambda_d \mathcal{L}_{\text{depth}} + \lambda_\tau \mathcal{L}_{\text{track}}$. The noise-adaptive confidence factor $c(\sigma)$ uses $\sigma_{\max}{=}0.85$ on the rectified-flow noise scale $\sigma\in[0,1]$, so the auxiliary losses are switched off only at the noisiest levels. The augmented track-state coefficients are $\alpha{=}0.5,\ \beta{=}0.25$. A linear warm-up of $1{,}000$ iterations ramps both auxiliary loss weights from $0$ to their full values to let the DiT backbone stabilize before the depth head dominates the gradient flow.

\textbf{Frozen modules.}
The WAN~2.1 VAE and the SigLIP2 encoder remain frozen throughout. Only the DiT parameters and the latent-space depth head (${\sim}138$K params) are updated.

\section{Two-Stage Inference Pipeline}
\label{sec:inference}

Our main-paper training procedure produces a fine-tuned diffusion prior over fixed-length 93-frame panoramic clips, conditioned on a perspective input video, SigLIP2 features, and a panoramic reference latent.
Producing the evaluation video from a single perspective input proceeds in two stages: a \emph{Round-1} pass that bootstraps a 93-frame ERP video using the same conditioning interface as training, and a \emph{Round-2} pass that consumes the first frame of Round~1 as an ERP image anchor and produces the final evaluation video via image-conditioned video2world.
All quantitative results in the main paper are computed on Round-2 outputs ($T{=}93$ frames at 16 fps, i.e.\ $\sim$5.8 seconds at $512 \times 1024$). Round~1 acts as an internal pre-generation step.

\textbf{Why a separate Round~2.}
A direct image-to-video call on the Round-1 first frame visibly degrades quality compared with Round-1 outputs.
The cause is that the fine-tuned DiT was trained to consume two extra inputs at every forward pass, the panoramic reference latent $\mathbf{z}_{\text{ref}}$ and the SigLIP2 cross-attention features $\mathbf{F}$.
Feeding zeros (the default behaviour of standard image-to-video calls) is severely out-of-distribution for the fine-tuned weights and produces washed-out colour, blurry texture, and reduced motion magnitude.

\textbf{Conditioning recipe for Round~2.}
We preserve the fine-tuning interface in Round~2 with three minor modifications relative to Round~1:
\begin{itemize}
\item The reference latent is the VAE-encoding of the entire Round-1 ERP video (93 frames), restoring the panorama-shaped context the DiT expects.
\item The Round-1 SigLIP2 features are reused verbatim, since they were derived from a perspective input that is in-distribution with training.
\item The spatial mask is set to one only on the first latent frame and zero elsewhere, indicating that Round~2 must trust the anchor frame and synthesize the remaining frames freely.
We do \emph{not} apply masked blending at every denoising step in Round~2 (in contrast to Round~1): blending would re-lock the frame-0 region to the ERP anchor at every step and suppress motion across the full sphere.
\end{itemize}

\textbf{Beyond the Round-2 horizon.}

The same procedure could be repeated to extend the horizon by re-encoding the previous round's ERP output as the new reference latent and taking its first frame as the new anchor.
We do not evaluate this longer-horizon setting in this work, and report all quantitative metrics on Round-2 outputs only.
Additional generated videos, including scenes with moving people and vehicles, are shown on the project page.

\section{Qualitative 3D Lifting of Generated Panoramas}
\label{sec:supple_3d_pipeline}

Because \pano{} is depth-supervised at training time, its panoramic output is already approximately geometry-consistent with a panoramic depth estimator, which lets us chain it with closed-form spherical lifting and an off-the-shelf rasterizer to lift a generated panorama into a coloured point cloud and render it from nearby viewpoints. We present this chain qualitatively. The pipeline has four stages.

\textbf{Step 1: Perspective image $\to$ panoramic video.}
Given a single perspective image (with a caption), the two-stage inference of Sec.~\ref{sec:inference} produces a $93$-frame ERP video at $512 \times 1024$ that captures the full $360^\circ$ surrounding scene with temporal motion.

\textbf{Step 2: ERP frame $\to$ panoramic depth.}
A panoramic depth map $\hat{D}_t \in [0, 1]^{H \times W}$ is recovered for each frame. Two depth sources are equally compatible: the latent depth head (${\sim}138$K params, Sec.~\ref{sec:losses}) used during training, which produces $\hat{D}_t$ in the same forward pass as the diffusion denoiser, or an off-the-shelf panoramic depth predictor such as DAP. Because the diffusion model is trained against DAP pseudo-GT, both choices stay geometrically consistent with the ERP frame at the same time index.

\textbf{Step 3: Spherical unprojection $\to$ dense 3D point cloud.}
Each ERP pixel $(u, v) \in [0, 1]^2$ together with its depth $\hat{D}_t(u, v)$ is lifted to a 3D Cartesian point via the same spherical unprojection $\pi_{\mathbb{S}^2}$ used by the trajectory loss (Eq.~\ref{eq:track_lift}). Stacking all valid pixels per frame yields a dense per-frame point cloud anchored in the world frame, and concatenating across the $93$ frames yields a temporally evolving 4D point cloud. A light planar regularizer collapses near-coplanar regions (walls, floor, ceiling) to suppress speckle from depth noise.

\textbf{Step 4: Rasterization $\to$ novel views.}
A virtual camera with pose $(\mathbf{R}, \mathbf{t})$ near the capture position and pinhole intrinsics $\mathbf{K}$ is rendered from the lifted scene. Direct point-based reprojection with $z$-buffering produces the novel views, and substituting per-point Gaussians and a 3D Gaussian Splatting rasterizer yields visually smoother novel-view videos at no extra training cost. Because the point cloud is lifted from a single viewpoint, regions occluded from the capture position are missing and appear as holes once the camera moves far from it.

\textbf{Open-source reference implementation.}
We release a stand-alone reference implementation of this pipeline. It reads a \pano{} ERP video, runs DAP on its first frame for stable per-pixel depth, applies the planar regularizer to produce a colored point cloud, renders novel-view trajectories (orbit, walk-through, fly-through) with a 3D Gaussian Splatting rasterizer, and finally upsamples the rendering with a 2x Real-ESRGAN super-resolution pass to compensate for residual blur near image boundaries. None of these post-processing modules are trained or fine-tuned for this work.

\textbf{Scope.}
From one perspective photograph and a caption, the chain yields (i) a $360^\circ$ panoramic video, (ii) per-frame panoramic depth, (iii) a 4D point cloud, and (iv) novel views near the capture position, at small additional cost on top of panoramic video generation. We do not evaluate the reconstructed geometry quantitatively, and the result inherits the errors of the monocular depth estimate; dedicated explorable-scene generators such as HunyuanWorld~1.0~\citep{hunyuanworld2025tencent} and One2Scene~\citep{wang2026one2scene} target this setting directly.

\section{Broader Impacts}
\label{sec:broader_impacts}

\textbf{Positive impacts.}
Geometry-consistent panoramic video generation lowers the production cost of immersive content for VR/AR education and training, virtual touring of inaccessible spaces (heritage sites, hazardous environments, remote locations), telepresence, and embodied-AI simulators that need full-sphere visual states from limited capture. The 3D lifting chain (Appendix~\ref{sec:supple_3d_pipeline}) may also lower the cost of simple 3D content creation from a single perspective photograph plus a caption.

\textbf{Potential negative impacts and mitigations.}
Like any high-fidelity generative video model, \pano{} could in principle be misused for immersive disinformation, fake virtual tours, or deepfake-style content embedded in $360^\circ$ media. We mitigate the most direct privacy risk by anonymizing all faces in the released self-captured corpus with Deface~\citep{xu2020centerface} (Sec.~\ref{sec:exp_setup}, Appendix~\ref{sec:supple_dataset}), and the released model checkpoints will ship with terms of use that prohibit deepfake / impersonation use, in line with the upstream Cosmos~Predict~2.5 safety policy that we inherit and do not bypass. Detection of synthetic panoramic content is an open problem; we encourage downstream users to combine our model with watermarking and provenance-tracking standards as they mature.

\end{document}